\documentclass[letterpaper]{article}
\usepackage{aaai}
\usepackage{times}
\usepackage{helvet}
\usepackage{courier}
\usepackage{todonotes}
\usepackage{amsmath}
\usepackage{amsthm}
\usepackage{times}
\usepackage{url}
\usepackage{graphicx}
\usepackage{subfig}
\usepackage{bbm}
\usepackage{amsfonts}
\usepackage{textcomp}
\usepackage{tabu}
\usepackage{multirow}
\usepackage{booktabs}
\usepackage{algorithm}
\usepackage[noend]{algpseudocode}
\usepackage[normalem]{ulem}
\usepackage{nicefrac}
\usepackage{csquotes}
\usepackage{enumitem}

\DeclareMathOperator*{\argmax}{arg\,max}

\begin{document}

\title{Learning to Rank Query Graphs for Complex Question Answering\\ over Knowledge Graphs}

\author{ Gaurav Maheshwari \\ Smart Data Analytics Group \\ University of Bonn, Germany
\And Priyansh Trivedi \\ Smart Data Analytics Group \\ University of Bonn, Germany
\And Denis Lukovnikov \\ Smart Data Analytics Group \\ University of Bonn, Germany 
\AND Nilesh Chakraborty \\ Smart Data Analytics Group \\ University of Bonn, Germany
\And Asja Fischer \\ Ruhr University, Bochum, Germany
\And Jens Lehmann \\ Smart Data Analytics Group \\ University of Bonn, Germany}

\maketitle

\begin{abstract}
\begin{quote}

In this paper, we conduct an empirical investigation of neural query graph ranking approaches for the task of complex question answering over knowledge graphs. 
We experiment with six different ranking models and propose a novel self-attention based \textit{slot matching} model which exploits the inherent structure of query graphs, our logical form of choice. 
Our proposed model generally outperforms the other models on two QA datasets over the DBpedia knowledge graph, evaluated in different settings. 
In addition, we show that transfer learning from the larger of those QA datasets to the smaller dataset yields substantial improvements, effectively offsetting the general lack of training data.

\end{quote}
\end{abstract}

\section{Introduction}
\label{sec:intro}

The increasing maturity of large-scale multi-relational knowledge graphs (KG) like DBpedia~\cite{lehmann2015dbpedia}, Freebase~\cite{bollacker2008freebase} and Wikidata~\cite{vrandevcic2014wikidata} has enabled a wide variety of interesting applications.
One of them is the problem of complex question answering over knowledge graphs (KGQA), 
which provides an intuitive natural language driven interface for accessing multi-relational knowledge.

Traditional KGQA approaches~\cite{dubey2016asknow,berant2014semantic} use semantic parsing to convert the natural language question (NLQ) to its corresponding formal query which is usually expressed in a formal language such as SPARQL or $\lambda$-DCS. This is done by
    (i) creating an expression representing the semantic structure of the question, and
    (ii) aligning this expression with the knowledge graph. 
While this approach suits the non-trivial task of handling wide syntactic and semantic variations of a question, it assumes the aforementioned tasks to be independent of each other.
This assumption can lead to undesirable situations where the system generates expressions which are illegal w.r.t.~the given KG.

In this work, we study an alternate family of approaches which treat the KGQA problem as that of generating a set of query paths on the KG and ranking them w.r.t.~the given question. 
These query graph ranking approaches tackle the aforementioned two steps in reverse order, i.e.~by
first constructing a list of candidate expressions w.r.t.~the KG schema, and 
then using the lexical and semantic structure of the NLQ 
to select the correct one.
This method ensures that all the candidate representations of the question correspond to the target KG structure. 
We use a custom grammar called \textit{query graphs} to 
represent these candidate expressions, comprised of paths in the KG along with some auxiliary constraints.

\begin{figure*}
\centering
\subfloat[{\scriptsize Question and Query Graph (partially correct)}]{
\centering
\includegraphics[width=0.33\textwidth,trim={0 1.5cm 0 0},clip]{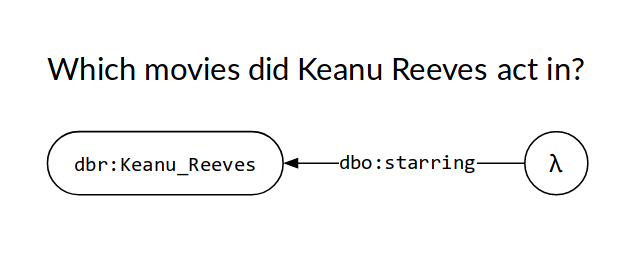}}
\subfloat[{\scriptsize Correct Query Graph}]{
\centering
\includegraphics[width=0.33\textwidth,trim={0 1.5cm 0 0},clip]{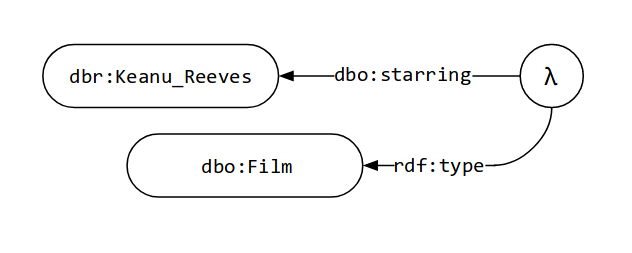}}
\subfloat[{\scriptsize Core chain corresponding to (b)}]{
\centering
\includegraphics[width=0.33\textwidth,trim={25cm 1.5cm 0 0},clip]{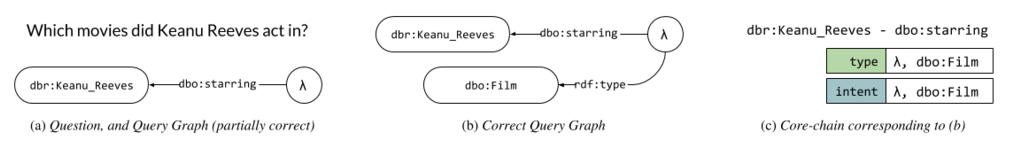}}
\caption{A question (a), its corresponding query graph (b), and core chain (c). Here the query graph in (a) is incomplete due to the absence of class constraint.}
\label{fig:querygraphs}
\end{figure*}

The primary objective of the study is to empirically investigate the effectiveness of query graph ranking approaches for the KGQA task. 
Motivated by the success of previous work in this direction \cite{yih2015semantic,bao2014knowledge},
we investigate
which ranking models are more suited for the task; what settings they are best trained in; what their limitations are; and explore further steps that can offset them.
To that end, we first evaluate simple models as  baselines and then appropriate existing models for our task.
We also propose a novel slot-matching model which exploits the structure of query graphs by comparing its parts with different representations of the question, computed using self attention. 
In our experiments, we find that it outperforms the other models.
Further, to illustrate the effects of various \emph{implicit decisions} that go into implementing this approach,  we evaluate these models in different settings, namely training in pointwise setting, pairwise setting, and with or without using shared parameters. 
The secondary objective of the study is to investigate whether transfer learning across similar tasks 
can increase the performance of the system.
We perform all our experiments over two KGQA datasets over DBpedia, namely, LC-QuAD~\cite{trivedi2017lc} and QALD-7~\cite{usbeck20177th}.

The primary contributions of this work are the following:
\begin{itemize}
    \item An evaluation of the effectiveness of numerous neural ranking models for ranking query graphs w.r.t.~NLQs.
    \item An investigation of the effect of transfer learning across two QA datasets. 
    \item A novel ranking model which exploits the characteristics of query graphs, and uses self attention and skip connections to explicitly compare each predicate in a query graph with the NLQ.
\end{itemize}

Through our experiments, we find that while the proposed slot matching model outperforms the others, a simple LSTM based encoder gives satisfactorily close results.
Moreover, our experiments reveal that pre-training models over LC-QuAD (larger dataset) and fine-tuning them over QALD-7 (smaller dataset) results in substantially better performance on the latter, as opposed to training them solely on QALD-7.

\section{Background}
\label{sec:background}

In order to formulate the problem, we first define the notion of a knowledge graph. 
Formally, let $\mathcal{E} = \{ e_1 \ldots e_{n_e} \}$ be the set of entities,  $\mathcal{L}$ be the set of all literal values, and $\mathcal{P} = \{p_1 \ldots p_{n_p} \}$ be the set of predicates connecting two entities, or an entity with a literal.
A knowledge graph $K$ is a subset of 
$(\mathcal{E} \times \mathcal{P} \times (\mathcal{E} \cup \mathcal{L}))$ representing the facts that are assumed to hold.

\subsection{Problem Formulation}
\label{sec:probdef}

Given a knowledge graph $K$, a natural language question $Q$, 
the system is expected to generate an expression of a
formal query language, 
which returns the intended answer $a \in A$ when executed over $K$. 
Here, $A$ is the set of all answers a KGQA system can be expected to retrieve, consisting of (i) a subset of entities ($e_i$) or literals ($l_i$) in $K$, (ii) the result of an arbitrary aggregation function ($f: \{e_i\} \cup \{l_i\} \mapsto \mathbb{N} $),
or (iii) a boolean ($T/F$) variable. 

\subsection{Query Graph}
\label{sec:querygraphs}

We use query graphs as the intermediary query language,
which represents paths in $K$  as a directed acyclic labeled graph.
We borrow the augmentations made to the language's grammar in~\cite{yih2015semantic}, which makes the conversion from query graph expressions to executable queries trivial.
We further make some changes to the grammar, as described below.

A query graph consists of a combination of nodes $n \in$ \{\textit{grounded entity}, \textit{existential variable}, \textit{lambda variable}, \textit{auxiliary function}\},  connected with labeled, directed edges representing the predicates ($p \in \mathcal{P}$). 
Each query graph has one or more \textit{grounded entities}, 
	which correspond to entities ($e \in \mathcal{E}$) present in the question $Q$.
The \textit{lambda variable} is not grounded to any entity in the KG, but instead acts as a placeholder (variable) for the set of entities which are the answer to the query, and some additional constraints (described below).
Similarly, the \textit{existential variables} aren't grounded, but are used to further disambiguate the structure of the query. 
Fig~\ref{fig:querygraphs}.a demonstrates that a question
can be represented with a lambda variable, and a grounded entity connected via a labeled edge.
Here the lambda variable can have many entities mapped to it, including $\mathsf{dbr}\mathsf{:}\mathsf{John\_Wick}$.

However, 
    this representation
	is incomplete, as the lambda variable can have other entities like a \textit{TV series}, or a \textit{play} mapped to it along with \textit{movies}, and thus its membership needs to be  constrained. 
These constraints can be enforced by the means of auxiliary functions, defined as follows: 
	(i) they can be of two types, namely,
    the cardinality function; and a class constraint 
    $ f_{class}:\{e \in \mathcal{E}~|~(e,~rdf\!:\!type,~class) \in K\}$ where \textit{(class, rdf:type, owl:Class)} $\in K$.
	(ii) these functions can only be applied on ungrounded nodes, i.e. \textit{lambda} and \textit{existential variables}. 
An updated query graph, with proper aggregation functions is denoted in Fig~\ref{fig:querygraphs}.b.

Finally, we define a new flag which determines whether the query graph is used to fetch the value of the projected variable, or to verify whether the graph is a valid subset of the target KG. 
The latter is used in the case of boolean queries like \textit{"Is Berlin the capital of Germany?"} We represent this decision with a flag instead of another node or constraint in the graph as it doesn't affect the execution of the query graph, but only inquires, post execution, whether the query had a solution.

\textbf{Representation:} 
We choose to represent the query graphs in a linear form so as to easily use them in our ranking models.
We linearize the directed graph by starting from one of the grounded entities and using $+,-$ signs to denote the outgoing and incoming edges, respectively. 
We further externalize the auxiliary functions, and their corresponding ungrounded nodes from the graph, and represent them with another flag along with the linearized chain.
Finally, we replace the URIs of entities and predicates with their corresponding surface forms. 
Hereafter, we refer to this linearized representation as the \textit{core chain} of a query graph.
This representation ensures that the query graph maintains textual relatedness to the source question, enabling us to use a wide variety of text similarity based approaches for ranking them.
Fig~\ref{fig:querygraphs}.c illustrates the core chain corresponding to the query graph in our running example.

\subsection{Scope}
\label{sec:scope}

We use the English version of DBpedia~\cite{lehmann2015dbpedia}, 2016-04 release,  as the KG for our question answering system.
It is a large scale KG consisting of 4.8M entities, and 580M triples\footnote{https://wiki.dbpedia.org/about}.

We restrict our system to answer
\emph{set} ($A_{set} \subseteq \mathcal{E} \bigcup \mathcal{L}$), 
\emph{simple count} ($A_{count}:= \{|a|: a \subseteq  \mathcal{E} \bigcup \mathcal{L}\})$
and 
\emph{boolean} queries ($A_{boolean} := \mathbf{1}_K(T)$
where $T,K \subseteq (\mathcal{E} \times \mathcal{P} \times (\mathcal{E} \cup \mathcal{L}))$,
$T$ is the set of triples from executing the query,
and $K$ is the KG).\footnote{$\mathbf{1}_A(\cdot)$ is the set indicator function.} Further, we restrict this study to queries which do not have more than two edges in the query graph, i.e. the shortest distance between the entity mentioned in $Q$, and the intended answer entity in $K$ is two.
These restrictions ensure a reasonably-sized candidate space, 
while maintaining enough expressivity to answer all questions in popular KGQA datasets like~\cite{trivedi2017lc,berant2013semantic}.

To prevent diluting the focus of our study, we assume entities $e^Q_1$ \ldots $e^Q_n$ to be given, as standalone entity-linking systems bring more uncertainty in the process and are not reflective of the inherent challenges in ranking query graphs.

\section{Approach}
\label{sec:approach}

The representation defined above enables us to divide the task of query graph construction and ranking in the following three phases: (i) \textit{core chain candidate generation}, (ii) \textit{core chain candidate ranking}, and (iii) \textit{predicting auxiliary constraints}.

\subsection{Core Chain Candidate Generation}
\label{sec:corechaingen}
This first step of the process involves generating core chain candidates for the question.
Core chains, as described in the previous section, are the linearized subset of the query graphs which represent a path consisting of entities and predicates without the additional constraints. 
Working under the assumption that the information required to answer the question is present in the target KG, 
	and that we know the entities mentioned in the question, 
	we collect all the plausible paths of up to two hops from the topic entity as the core chain candidate set.
Here, the number of hops of a core chain equals the number of predicates in the core chain and the $n^{th}$ hop in a core chain is the $n^{th}$ predicate in the core chain, counting from a grounded node.

We retrieve candidate core chains by collecting all predicates (one-hop chains) and paths of two predicates (two-hop chains) that can be followed from an arbitrary grounded node\footnote{Entity that has been linked in the question.}.
In this process, predicates are followed in both forward and reverse direction (and marked with a $+$ and $-$ in the chain, respectively). 
For LC-QuaD, we also try to restrict our candidate set of relational chains as follows: if two entities have been identified in the question, only two hop chains are retained that connect the first grounded entity with the second, leaving the answer node in between. 
Finally, we reject the core chains where all the entities linked in the question aren't satisfied by its corresponding query graphs. In cases where there are multiple entities, this step substantially decreases the candidate set while retaining all the relevant groundings.

Although we limit the core chains to a length of two hops for the purposes of this study, this approach can be generalized for longer core chains, however it may result in additional challenges because of the exponential explosion of candidate core chains. 

\subsection{Core Chain Candidate Ranking}
\label{sec:ranking}
After generating a set of core chain candidates for a given question, we employ a neural ranking model to select the most plausible core chain. 
This is done by computing the similarity of a core chain with the input question. 
Given a question utterance $Q = \left[q_0 \ldots q_T\right]$ where $q_i$ is the $i^{th}$ word in the question, and similarly a core chain $C = \left[c_0 \ldots c_{T^\prime}\right]$, we model the scalar similarity score as follows:
{\begin{equation}
    sim(Q, C) = com\big(enc^q(Q), enc^c(C)\big) \enspace ,
\end{equation}
where the encoder functions, ($enc^q, enc^c$)
and the compare function~($com$)~are trained jointly. After generating a set of candidate core chains, $C_0 \ldots C_n$, we select the most plausible core chain as follows:
\begin{equation}
    C^* = \argmax_{C_i} sim(Q, C_i)
\end{equation}
We studied several encoding and comparing functions, trained in multiple configurations, as described below.

\subsubsection{Encoders}
\label{sec:encoders}
Encoders translate a core chain or a question into a fixed-length vector representation.

We include some simple models as the baselines for our study, propose a novel \emph{slot matching} encoder, and appropriate two existing models~\cite{parikh2016decomposable,yu2017improved}, which we find suitable for our task. 

\subsubsection{Bidirectional LSTM:} We use Bidirectional Long-Short Term Memory\cite{hochreiter1997long} networks (BiLSTM) for encoding both the question and the core chains.
The inputs $Q$ and $C$ are
treated sequences of words 
which are embedded ($\mathsf{EMB(\cdot)}$) using pre-trained GloVe~\cite{pennington2014glove} vectors, and are passed to the LSTM layers:

\label{para:bilstm}
\begin{align}
    \label{eq:bilstm}
    \vec{q} &= enc^q_{\mathsf{LSTM}}(Q) := \mathsf{BiLSTM_q}(\mathsf{EMB}(Q))\\
    \vec{c} &= enc^c_{\mathsf{LSTM}}(C) := \mathsf{BiLSTM_c}(\mathsf{EMB}(C)) \enspace 
\end{align}

where the functions $\mathsf{BiLSTM(\cdot)}$ return the final hidden state of the LSTM. 

\subsubsection{CNN:}
\label{para:cnn}
We also explore using CNN architecture, as proposed in~\cite{kim2014convolutional}, which uses three multi-channel convolutions to create multiple feature vectors of a sequence, which are then max-pooled across time, concatenated, and passed through a linear layer to create its fixed-length representation.
They are modelled analogous to the BiLSTM encoders and represented as $enc_{\mathsf{CNN}}^q$ and $enc_{\mathsf{CNN}}^c$, respectively.

\subsubsection{Slot Matching Model (Novel):}

The previously described models encode both the question and core chain each into their respective vector representations, which forces complex transformations on both sequences, possibly introducing a bottleneck in optimal performance.
In order to alleviate this problem, we propose a more structured encoding scheme which partitions the core chain into hops, and creates multiple representations of the question called \textit{slots} corresponding to each hop, which are then individually compared.

The proposed model works as follows.
First, the question is encoded using a BiLSTM. For the $j^{th}$ hop in the core chain, we define a trainable slot attention vector $k_j$ that is used as the query vector to compute attention weights $\alpha_{t,j}$ over $Q = \left[q_0 \ldots q_T\right]$.
Note that $k_j$ is shared across all examples.
Finally, a slot-specific question representation $\vec{q}_j$ is computed, by first adding the word vectors
to the encoding, and using the corresponding scalar attention weights $\alpha_{t,j}$ to summarize over time.
This process can be summarized as follows:
\begin{align}
	\left[\mathbf{q}_0 \dots \mathbf{q}_T\right] &= enc_{\mathsf{LSTM}}^q(Q) \label{returnall} \enspace \\
	\alpha_{t,j} &= softmax(\{\mathbf{q}_l \cdot k_j\}_{l=0\ldots T})_t \enspace \\
	\vec{q}_j &= (enc^q_{\mathsf{SLOT}}(Q))_j :=  \sum^{T}_{t=0}~\alpha_{t,j}(\mathsf{EMB}(q_{t}) + \mathbf{q}_t) \enspace .
\end{align}

We represent the core chains by separately encoding each hop (directions and predicate's surface form) by another LSTM ($enc^c_{\mathsf{LSTM}}$), and add skip connections from the embedding layer to it:

\begin{align}
	\vec{c}_{j} &= enc^c_{\mathsf{LSTM}}(C^j) + \frac{1}{T_j^\prime}\sum^{T_j^\prime}_{t=0}\mathsf{EMB}(c^j_t) \enspace ,
	\end{align}
where $C^j = [c^j_0 \ldots c^j_{T_j^\prime}]$ is the sequence of words in the $j^{th}$ hop of the core chain.
Finally, $\vec{q}_j$ and $\vec{c}_j$ for the different slots $j$, as just defined, are concatenated to yield $\vec{q} 
= enc^q_{\mathsf{SLOT}}(q_{0\ldots t}) := [ \vec{q}_1, \vec{q}_2 ] $ 
and $\vec{c}
= enc^c_{\mathsf{SLOT}}(c_{0\ldots t^\prime}) := [ \vec{c}_1, \vec{c}_2 ] $

Note that the model proposed here deviates from  \emph{cross attention} between the input sequences (which we also experiment with, as described below) as, in our case the attention weights aren't affected by the predicates in the core chain, as the encoder attempts to focus on \emph{where} a predicate is mentioned in $Q$, and not \emph{which} predicate is mentioned. 

\subsubsection{Decomposable Attention Model:}
\cite{parikh2016decomposable} proposes a novel model for the natural language inferencing task, which computes a summary of two input sequences, weighted by soft cross-attention, signaling the important parts of a sequence w.r.t others.
We hypothesize that this local alignment based model, defined as follows, while susceptible to overfitting (due to increased model complexity), might be effective in our use case.

We use the aforementioned BiLSTM layer to encode the question and the core chain, and use the \textit{attend-align-compare} layers to compute a fixed-length summary vector for both sequences. Then, we use skip connections and concatenate the last states of the encoded vector and the summarized vectors. 

\begin{align}
    \vec{q},\ \vec{c} &= enc_{\mathsf{DAM}}(Q, C) := \mathsf{DAM}(enc_{LSTM}^q(Q), enc_{LSTM}^c(C))
\end{align}

Here $\mathsf{DAM}(\cdot,\cdot)$ is the function which calculates the summary vector from the model proposed in~\cite{parikh2016decomposable}. Notationally, this encoder, $enc_{\mathsf{DAM}}$ is slightly unlike the rest, as it requires both sequences $Q$ and $C$ as inputs to compute alignments of one sequence against the other. 

\subsubsection{Hierarchical Residual Sequence Model:} Finally, we appropriate another neural ranking model, originally proposed in~\cite{yu2017improved} for a task closely aligned to ours, namely, that of ranking KG predicates for relation detection in questions. 
The model learns to create a hierarchical representation of the question, and a representation of a set of predicates across different granularities (word level, relation level):

\begin{align}
    \vec{q} =& \frac{1}{2}\big(enc^q_{\mathsf{LSTM}}(Q) + \mathsf{BiLSTM_{q2}}(enc^c_{\mathsf{LSTM}}(Q)) \big)\\
    \vec{c} =& \frac{1}{2}\big(\frac{1}{l_{cw}}\sum^{l_{cw}}(\mathsf{BiLSTM}_{cw}(C)) +                                  
      \frac{1}{l_{cp}}\sum^{l_{cp}}(\mathsf{BiLSTM}_{cp}(C^{\mathsf{pred}}))\big)
      \label{eq:rhs:c}
\end{align}

\noindent where $\mathsf{BiLSTM}_{cw}$ encodes the core chains as a sequence of words $C$
, $\mathsf{BiLSTM}_{cp}$ encodes the core chains as a sequence of predicates $c^{\mathsf{pred}}$. Like the other encoders, we define $\vec{q} = enc^q_{\mathsf{HRM}}(Q)$ and $\vec{c} = enc^c_{\mathsf{HRM}}(C)$

\paragraph{Compare Function:}
\label{sec:compare}
We use a function $com(\cdot)$ which computes a single scalar from the vector representations of the question and the core chain respectively. 
To compute $com(\cdot)$ in our experiments, we use (i) the dot product, (ii) a feed-forward layer to encode question and core chain representations and then compute the dot product: 
\begin{align}
    com_{\mathsf{dot}}(\vec{q}, \vec{c}) &= \vec{q} \cdot \vec{c} \\
    com_{\mathsf{dense\_dot}}(\vec{q}, \vec{c}) &= \mathrm{FF}(\vec{q}) \cdot \mathrm{FF}(\vec{c}) 
\end{align}
where $\mathrm{FF}$ is a feed-forward layer. 
The $sim(\cdot)$ function, defined above, is thus composed using one of the five encoders, and one of these comparison functions. 

\subsection{Predicting Auxiliary Constraints}
\label{sec:auxconstraints}
In this phase, we learn
to predict 
the auxiliary constraints and flags used for constructing a complete query graph.
We begin by predicting the intent of the question.
In both the datasets in our experiments, a question can ask for the cardinality of the projected variable, ask whether a certain fact exists in the KG, or simply ask for the set of values in the projected variable. 
Further, this division, hereafter referred to as \textit{count}, \textit{ask} and \textit{set} based questions, is mutually exclusive.
We thus use a simple BiLSTM based classifier to predict the intent as one of the three.

Next, we focus on detecting class based constraints on the ungrounded nodes of the core chain. 
For instance, in the following question: \textit{"Which movies has Keanu Reeves starred in?"}, the word \textit{movies} constraints the list of everything that Keanu Reeves starred in , including tv shows, plays, etc.
In SPARQL, these constraints are expressed as a triple like
$(\mathsf{?x}~\mathsf{rdf}\mathsf{:}\mathsf{type}~\mathsf{dbo}\mathsf{:}\mathsf{className})$.
We represent them in our query graphs as auxiliary constraints on either the existential or lambda variable.
We use two different, separately trained models to predict (i) whether such a constraint exists in the question, and if so, on which variable, and (ii) which class is used as a constraint.
The former is accomplished with a simple BiLSTM, akin to the aforementioned intent classifier.
For the latter, we use a pairwise ranking based model, specifically the first model mentioned in the core chain candidate ranking section. 
Further details of training all these models can be found in Approach Evalutaion section.

We now have all the information required to construct the query graph, and the corresponding SPARQL. 
For brevity's sake, we omit the algorithm to convert query graphs to SPARQL here, but for limited use cases, simple template matching shall suffice.

\section{Experiments}

In this section, we describe the different experiments we perform, and their results.
\begin{table*}[h!]
\centering
\subfloat[Performance on LC-QuaD. \label{tab:results:lcquad}]{\centering
\begin{tabular}{@{}r|ccccc|ccccc@{}}
\toprule
                        \multicolumn{1}{c}{}    & \multicolumn{10}{c}{LC-QuAD}                                                                                                                            \\ 
\midrule
 & \multicolumn{5}{c|}{Pointwise} & \multicolumn{5}{c}{Pairwise}\\
 & CCA   & MRR   & P     & R    & F1 & CCA & MRR   & P    & R    & F1            \\
\midrule
BiLSTM Dot & 0.56 & 0.64 & 0.63 & 0.74 & 0.68 & 0.53 & 0.62 & 0.59 & 0.71 & 0.64          \\
BiLSTM Dense Dot           &    0.42   & 0.44  &    0.50   &    0.61  & 0.55 & 0.45      &  0.55     & 0.53     & 0.65     &       0.58        \\
CNN Dot                    &     0.37  &    0.47   &    0.45   &    0.57  & 0.50 &  0.41     & 0.51 &  0.50    &    0.61  &      0.55         \\
DAM Dot                    & 0.48 & 0.58                     & 0.57  & 0.68 & 0.62          & 0.50 & 0.59                     & 0.58 & 0.69 & 0.63          \\
HRM Dot                    &    0.54   &    0.64                       &    0.62   &    0.73  &   0.67 &    0.47   &    0.57 & 0.55     & 0.67     &    0.60           \\
\textbf{Slot-Matching Dot} & 0.58  & 0.66                      & 0.65  & 0.76 & \textbf{0.70} & 0.58  & 0.66                      & 0.66 & 0.77 & \textbf{0.71} \\ \bottomrule
\end{tabular}
}

\subfloat[Performance on QALD-7.\label{tab:results:qald}]{\centering
\begin{tabular}{@{}r|ccccc|ccccc@{}}
\toprule
\multicolumn{1}{c}{}     & \multicolumn{10}{c}{QALD-7} \\ \midrule
 & \multicolumn{5}{c|}{Pointwise} & \multicolumn{5}{c}{Pairwise} \\
         & CCA  & MRR  & P    & R    & F1   & CCA  & MRR  & P    & R    & F1   \\
\midrule
BiLSTM Dot       & 0.30 & 0.39 & 0.25 & 0.41 & 0.31 & 0.25 & 0.37 & 0.20 & 0.35 & 0.25 \\
BiLSTM Dense Dot &  0.41    & 0.26    & 0.26     & 0.38     & 0.31 &    0.28  & 0.37     & 0.22     & 0.38     &   0.28   \\
CNN Dot          & 0.23 & 0.29 & 0.12 & 0.28 & 0.17 & 0.25 & 0.32 & 0.20 & 0.35 & 0.25 \\
\textbf{DAM Dot}          & 0.35 & 0.47                     & 0.28 & 0.43 & \textbf{0.34} & 0.25 & 0.37                     & 0.22 & 0.38 & \textbf{0.28} \\
HRM Dot          &   0.31   &    0.41 & 0.26     &  0.43     &  0.32 &  0.20    &    0.32                      & 0.20     &    0.36  &    0.26  \\
Slot-Matching Dot                     & 0.33 & 0.43                     & 0.22 & 0.38 & 0.28                      & 0.25 & 0.40                     & 0.17 & 0.33 & 0.22 \\ 
\bottomrule
\end{tabular}
}
\caption{Performance on LC-Quad and QALD-7. The reported metrics are core chain accuracy (CCA), mean reciprocal rank (MRR) of the core chain rankings, as well as precision (P), recall (R) and the F1 of the execution results of the whole system.
\label{tab:results}}
\end{table*}

\subsection{Approach Evaluation}
\label{sec:expmain}

Our first experiment focuses on the accuracy of different ranking models as discussed in the previous section, and their effect on the overall performance of the pipeline. 

\subsubsection{Datasets}
\label{sec:expmain:data}
We use the LC-QuAD~\cite{trivedi2017lc} and QALD-7-multilingual~\cite{usbeck20177th} datasets to evaluate the performance of our system.

\textbf{LC-QuAD} is a gold standard question answering dataset over the DBpedia 04-2016 release, having 5000 NLQ and SPARQL pairs.
The coverage of our grammar (as defined in the Background section) covers all kinds of questions in this dataset. 

\textbf{QALD} is a long running challenge for KGQA over DBpedia. 
While currently its $8^{th}$ version is available, we use QALD-7 (Multilingual) for our purposes, as it is based on the same DBpedia release as that of LC-QuAD.
QALD-7 is a gold-standard dataset having 220 and 43 training and test questions respectively along with their corresponding SPARQL queries. 
This dataset is more diverse than LC-QuAD, with some of the questions are outside the scope of our system.
We nonetheless consider all the questions in our evaluation.

\subsubsection{Evaluation Metrics}
We measure the performance of the proposed methods in terms of its ability to find the correct core chain, as well as the execution results of the whole system.
For core chain ranking, we report Core Chain Accuracy (CCA) and Mean Reciprocal Rank (MRR).
Based on the execution results of the whole system (including auxiliary components), we also report Precision, Recall and F1.

\subsubsection{Training}
\label{sec:exp1:training}

We train our core chain ranking models both in pointwise and pairwise setting, using negative log likelihood, and max-margin losses respectively. Both are commonly used in ranking problems.
In both settings, our models are trained with negative sampling, where we sample 100 negative core chains per question, along with the correct one for every iteration.
In the pointwise training setting, the objective is to maximize the score of the correct core chain and minimize the score of the negative samples:
$\mathcal{L}_{point} = - (t \cdot \log(s) + (1 - t) \cdot \log(1 - s))$, 
where $t$ is 1 if the question-chain pair is correct and 0 otherwise.
In contrast, in pairwise training, we maximize the difference of the scores, up to some margin: $\mathcal{L}_{pair} = \max (0, \gamma - s^+ + s^-)$,
where 
$s^+$ and $s^-$ are the scores for correct and incorrect question-chain pairs, respectively.

Our models are trained for 300 epochs, with early stopping enabled based on validation accuracy.
We use a 70-10-20 split as train, validation and test data over LC-QuAD\footnote{in keeping with the train-test splits suggested by the authors.}. 
QALD-7 has a predefined train-test split, we however use one eighth of the train data for validation. 
In both cases, we do not train on the validation data. 
We embed the tokens using Glove embeddings\footnote{Trained over the common Crawl corpus. 300 dimensions, and with 1.9M tokens in the vocabulary.}~\cite{pennington2014glove}, 
and keep the relevant subset of the embeddings trainable in the model. 
We use the Adam optimizer, set the learning rate to 0.001, and clip gradients at 0.5.
In this experiment, we share parameters between $enc^c$ and $enc^q$ for the BiLSTM and CNN models which increases their performance. 
We discuss the effects of parameter sharing in another experiment below.

The intent and rdf-type existence prediction models are trained without negative samples, but with early stopping with the same splits, and the rdf-type class prediction model is trained akin to the core chain ranking models, with negative sampling and early stopping.
These models are not trained jointly, which we intend to explore in the future.

\subsubsection{Results and Error Analysis}
\label{sec:exp1:res}

In our experiments, as detailed in Table~\ref{tab:results}, we observe that the F1 score of almost all models is within a short range of 60\% to 71\%. 
The slot matching model performs the best among them, both in pointwise and pairwise settings.
Upon closer inspection, we find that the model learns to attend over the entities and predicates in the question as visualized in Fig~\ref{fig:attn}. 
While the DAM dot also uses attention, its performance generally lags behind the slot matching model. 
We attribute this to the fact that the DAM dot model tries to create a new representation of the question for each core chain (due to cross-attention between core-chain and question sequences);  while a question's representation does not depend on the corresponding core chain in the slot matching model, thereby helping it generalize better.
The performance of the $\mathsf{BiLSTM}$ model with $\mathsf{dot}$ encoder is in keeping with recent findings in~\cite{mohammed2017strong}, i.e. a simple  recurrent model can perform almost as well as the best performing alternative. 

Overfitting is generally observed across all our models trained over LC-QuAD, and is much worse in the case of QALD-7. 
For instance, the slot matching model, trained over LC-QuAD in pairwise setting has a core chain accuracy of 93.14\% over the training data. 
All our models peformed poorly on QALD-7, with the best being DAM dot.
This isn't surprising given the fact that QALD-7 has only 220 examples in the training set,  which is 20 times smaller than LC-QuAD.
We will show in the next section that transfer learning across datasets is a viable strategy in this case to improve model performance.

\begin{table}[h!]
\centering
\begin{tabular}{@{}lcccc@{}}
\toprule
                                      & \multicolumn{2}{c}{Pointwise} & \multicolumn{2}{c}{Pairwise} \\
                                      & (i)           & (ii)          & (i)           & (ii)         \\ \midrule
\multicolumn{1}{r}{BiLSTM Dot}        & 0.49            & 0.36         & \textbf{0.51}          & 0.31         \\
\multicolumn{1}{r}{Slot Matching Dot} & 0.49          & 0.41          & \textbf{0.54}          & 0.25         \\ \bottomrule
\end{tabular}
\caption{CCA for Transfer Learning Experiments.}
\label{tab:transfer}
\end{table}

\begin{figure}[]
  \centering
  \includegraphics[width=0.475\textwidth,trim={0.24cm 0 0 0},clip]{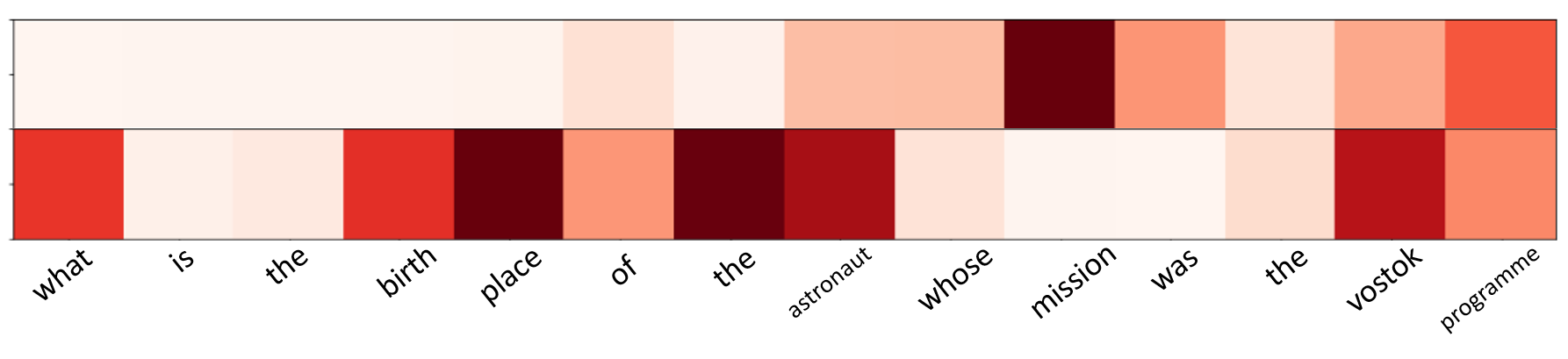}
  \caption{Visualized attention weights of the slot matching question encoder for the  question "\textit{What is the birth place of the astronaut whose mission was the vostok programme?}" Here, the first row represents $k_1$ and second $k_2$}
  \label{fig:attn}
\end{figure}

\subsection{Transfer Learning}
\label{sec:exptransfer}

As mentioned above, neural ranking models seem to be ineffective when training solely on QALD-7 due to a more varied and noticeably smaller dataset. We hypothesize that using LC-QuAD to pretrain the ranking models might lead to a significant increase in performance.

\begin{itemize}[leftmargin=2em]

	\item[(i)] We pre-train our ranking models over LC-QuAD, fine-tune them over QALD-7's train split, and evaluate over its test split.

	\item[(ii)] We also perform this experiment in a simpler setting, by coalescing the training data across the two datasets and testing over QALD-7.

\end{itemize}

Both the experiments are conducted over the two best performing encoders: $enc_{\mathsf{LSTM}}, enc_{\mathsf{SLOT}}$, and using $com_{\mathsf{dot}}$. 
The results of this experiment are mentioned in Table~\ref{tab:transfer}. 
Here, we only report the core chain accuracies as the rest of the system remains unchanged for the purposes of this experiment. 

In our experiments, we find that while both forms of transfer learning improve the performance of the ranking models, fine-tuning is more effective in every setting.
While the slot-matching model trained in pairwise setting and fine-tuned for QALD-7 gives the best performance, we observe that pointwise models exhibit a more consistent performance improvement. 
While fine-tuning our models, the initial learning rate was set to 0.0001, i.e. an order of magnitude less than in the first experiment. 
Not doing so leads to the models overfitting on the training dataset. For instance, the slot matching model fine-tuned in a pairwise setting with 0.001 learning rate did not show any improvement. 

\subsection{Further Analysis}
\label{sec:anal}
In order to better assess the impact of different parts of the system, we perform a series of analyses over our two best performing models, i.e. slot matching dot and BiLSTM dot. 
Also, we only calculate the core chain accuracy in this case 
unless specified otherwise, and the hyperparameters for each experiment are the same as mentioned in the first experiment.

\subsubsection{Pointwise vs Pairwise}
As observed in Table~\ref{tab:results:qald}, and~\ref{tab:transfer}, pointwise models generally outperform their pairwise counterparts when trained on small datasets but have a comparable performance otherwise.
To analyze whether the performance gain in anecdotal, specific to QALD-7, or can be generalized to smaller datasets, we train our models on one fifth of the LC-QuAD's training split. 
We find that across multiple runs, BiLSTM dot trained in a pointwise setting consistently outperforms its pairwise counterpart (34\% and 31\% respectively), and the vice versa holds for the slot matching model (31\% pointwise, 36\% pairwise).
Thus, the results are inconclusive, indicating the fact that performance gain in QALD-7 is due to the innate characteristics of the dataset.

\subsubsection{Parameter Sharing} Given that in the first experiment, the BiLSTM dot model shares parameters across the encoders and the slot matching model doesn't, i.e., $enc_{\mathsf{LSTM}}^q = enc_{\mathsf{LSTM}}^c$, $enc_{\mathsf{SLOT}}^q \neq enc_{\mathsf{SLOT}}^c$, we intend to find the effect of sharing encoder parameters. To do so, we experiment with both the slot matching dot and BiLSTM dot models, with and without sharing encoder parameters.

We find that sharing parameters between encoders for the slot matching dot model, trained in a pairwise setting results in performance improvement (+2\%), thereby making it the best performing model across all settings. 
However, in a pointwise setting, the performance drops by 6\% instead. 
Similar results are observed when we use use this (pointwise) model in the experimental setup of the second experiment.
On the other hand, not sharing parameters across the encoders in BiLSTM dot results in a consistent performance drop (-6\%). We attribute this to the siamese nature of BiLSTM dot model, i.e. both encoders process their inputs in the same manner, whereas in the slot matching dot model, $enc^q_{\mathsf{SLOT}}$ uses self attention to create different representations of the inputs whereas $enc^c_{\mathsf{SLOT}}$ is a simpler recurrent encoder with residual connections. 

\subsubsection{Auxiliary Component Analysis}
In this subsection we discuss the performance of the auxiliary components of our system, namely, intent prediction, rdf-type existence, and rdf-type class prediction models.
The intent prediction model solves a relatively easier task of sequence classification between {$set, count$ and $ask$}. This model gives the test accuracy of 99.1\% and 91.8\% when trained over LC-QuAD and QALD-7 respectively.
The rdf-type existence model performs a similar task of predicting whether a class constraint is implied in the question, and if so, on which variable. It performs with the accuracy of 75.3\% over LC-QuAD. It's performance over QALD-7 is as less as 37.2\%, due to disproportionately weighted example distribution. Due to this reason, we simply use a pre-trained model trained over LC-QuAD for our purposes. In our experiments, it performs with a 77.0\% accuracy. 
The same holds for the rdf-type class prediction model which predicts the $\mathsf{owl:Class}$ for the class constraint, if applicable. This model performs with 69\% on LC-QuAD. 

\section{Related Work}
\label{sec:relwork}
The state-of-the-art methods for complex QA over knowledge graphs take primarily three different kinds of approaches - 
(i) using semantic parsers to create NLQ representations which are then grounded against the KG,
(ii) generating grounded candidate representations for NLQ and re-ranking question-graph pairs, 
(iii) neural sequence decoding models that directly generate a logical form from a given NLQ. 
We keep our discussion of related literature restricted to the first two. 

Traditional semantic parsing based KGQA approaches~\cite{dubey2016asknow,xu2014answering,fader2014open,berant2014semantic,reddy2017universal,cui2017KBQA}
aim to learn semantic parsers that generate \textit{ungrounded} logical form expressions from NLQs, and subsequently ground the expressions semantically by querying the KG. 

In recent years, several papers have taken an alternate approach to semantic parsing by treating it as a problem of semantic graph generation and re-ranking.
\cite{bast2015more} compare a set of manually defined query templates against the NLQ and generate a set of grounded query graph candidates by enriching the templates with potential predicates.
Notably,~\cite{yih2015semantic} creates grounded query graph candidates using a staged heuristic search algorithm, and employ a neural ranking model for scoring and finding the optimal semantic graph. 
The approach we propose in this work is closely related to this.
\cite{yu2017improved} use a hierarchical representation of KG predicates in their neural query graph ranking model.
They compare their results against a local sub-sequence alignment model with cross-attention~\cite{parikh2016decomposable}. 
We appropriate the models proposed by both~\cite{parikh2016decomposable} and~\cite{yu2017improved} for our task, and compare against the baselines we propose.

\section{Conclusion and Future Work}
\label{sec:conc}
We studied the performance of neural ranking models for ranking query graphs
	and showed that this family of approaches can be used for question answering against a KG.
We further explored the effects of numerous variations in structure, training, and hyperparameters, while evaluating the models on LC-QuAD and QALD-7.
We also proposed a novel task specific ranking model which outperforms the others, in our experiments.
Finally, we showed that transfer learning can be effective to offset the lack of training data.

We aim to extend this work by using techniques to transfer knowledge from pre-trained language models, trained over domain agnostic text. 
Further, we intend to explore mechanisms enabling in-network answer supervision based on differentiable query execution~\cite{cohen2016tensorlog}. 
\newpage
\bibliographystyle{aaai}

\end{document}